\documentclass[conference]{IEEEtran}
\IEEEoverridecommandlockouts

\usepackage[OT1]{fontenc} 

\usepackage{cite}
\usepackage{amsmath,amssymb,amsfonts}
\usepackage{algorithm}
\usepackage{algorithmic}
\usepackage{graphicx}
\usepackage{textcomp}
\usepackage{xcolor}
\usepackage{multirow}

\usepackage{booktabs}
\usepackage{tabularx}
\usepackage{pgf-pie}
\usepackage{pgfplots}
\pgfplotsset{compat=1.18}
\usepackage{url}
\usepackage{hyperref}
\hypersetup{
    colorlinks=true,
    linkcolor=black,  
    citecolor=blue,   
    urlcolor=blue,    
    pdfauthor={Mohamed Basem, Islam Oshallah, Ali Hamdi, Khaled Shaban, Hozaifa Kassab},
    pdftitle={Two-Stage Quranic QA via Ensemble Retrieval and Instruction-Tuned Answer Extraction}
}

\def\BibTeX{{\rm B\kern-.05em{\sc i\kern-.025em b}\kern-.08em
    T\kern-.1667em\lower.7ex\hbox{E}\kern-.125emX}}

\begin{document}

\title{Two-Stage Quranic QA via Ensemble Retrieval and Instruction-Tuned Answer Extraction
}
\author{
Mohamed Basem$^{1}$, Islam Oshallah$^{1}$, Ali Hamdi$^{1}$, Khaled Shaban$^{2}$, Hozaifa Kassab$^{3}$ \\
$^{1}$Dept. of Computer Science, MSA University, Giza, Egypt \\
\{mohamed.basem1, islam.abdulhakeem, ahamdi\}@msa.edu.eg \\ 
$^{2}$Dept. of Computer Science, Qatar University, Doha, Qatar \\
 khaled.shaban@qu.edu.qa \\
$^{3}$AiTech AU, Australia \\
hozaifa@aitech.net.au
}

\maketitle

\begin{abstract}
Quranic Question Answering presents unique challenges due to the linguistic complexity of Classical Arabic and the semantic richness of religious texts. In this paper, we propose a novel two-stage framework that addresses both passage retrieval and answer extraction. For passage retrieval, we ensemble fine-tuned Arabic language models to achieve superior ranking performance. For answer extraction, we employ instruction-tuned large language models with few-shot prompting to overcome the limitations of fine-tuning on small datasets. Our approach achieves state-of-the-art results on the Quran QA 2023 Shared Task, with a MAP@10 of 0.3128 and MRR@10 of 0.5763 for retrieval, and a pAP@10 of 0.669 for extraction, substantially outperforming previous methods. These results demonstrate that combining model ensembling and instruction-tuned language models effectively addresses the challenges of low-resource question answering in specialized domains.

\end{abstract}

\vspace{3mm}


\section{Introduction}

The Holy Qur'an, revealed over 1,400 years ago, remains the primary source of guidance for over 1.8 billion Muslims worldwide. Beyond its religious significance, the Qur'an represents a masterpiece of Classical Arabic literature, containing profound linguistic, historical, and ethical insights that continue to be studied by scholars across multiple disciplines~\cite{mattson2012story}. However, accessing and understanding specific Quranic content in response to natural language queries presents significant computational challenges that have only recently begun to be addressed by the natural language processing community~\cite{utomo2020question ,BadaroSurvey}.

Question Answering (QA) systems have emerged as a cornerstone technology for information access across diverse domains, from general knowledge retrieval to specialized applications in legal, medical, and educational contexts~\cite{martinez2023survey,wang2024large}. The development of effective QA systems for religious texts, particularly the Qur'an, represents both a meaningful application with substantial social impact and a challenging technical problem that pushes the boundaries of current Arabic NLP capabilities~\cite{utomo2020question ,ReviewChallenging}.

Quranic Question Answering presents unique challenges that distinguish it from conventional QA tasks. First, the linguistic complexity of Classical Arabic—with its rich morphology, complex syntax, and archaic vocabulary—poses significant difficulties for models primarily trained on Modern Standard Arabic or contemporary dialects~\cite{altammami2022challenging}. Second, the semantic richness and multi-layered meanings inherent in Quranic verses require sophisticated understanding that goes beyond surface-level text matching. Third, the limited availability of large-scale, high-quality annotated datasets for Quranic content constrains the development and evaluation of machine learning models~\cite{malhas2020ayatec,quqa2023,abdallah2024arabicaqa}.
Finally, the cultural and religious sensitivity of the domain requires both precision and humility. Quranic interpretation has traditionally been reserved for scholars, and AI outputs risk being mistaken for authoritative tafsīr, potentially misleading users or fueling sectarian disputes. To avoid this, systems should present answers as informational aids rather than interpretive rulings~\cite{nasr2015studyquran}.

Recent advances in transformer-based language models have shown promising results for Arabic NLP tasks~\cite{antoun2020arabert,antoun-etal-2021-araelectra , MohamedMulti , AllamArabic}. However, these models face particular challenges when applied to Classical Arabic texts, where the linguistic gap between training data and target domain significantly impacts performance~\cite{inoue-etal-2021-interplay}. While ensemble methods have proven effective in improving robustness across various NLP tasks~\cite{mrqa2019}, their application to low-resource, high-complexity domains like Quranic QA remains underexplored~\cite{elkomy2022}. Similarly, the emergence of instruction-tuned large language models offers new possibilities for few-shot learning in specialized domains, yet their effectiveness for Classical Arabic text understanding has not been thoroughly investigated~\cite{zhu2023large,ramachandran2024gemini,liu2024deepseek}.

The Qur'an QA 2023 Shared Task was designed to address these challenges through two complementary tasks: (1) Passage Retrieval, which requires identifying the most relevant Quranic verses for a given natural language question, and (2) Machine Reading Comprehension (MRC), which focuses on extracting precise answer spans from retrieved passages~\cite{malhas-etal-2023-quran}.  Despite the participation of multiple research teams ~\cite{elkomy2023tce,alawwad2023ahjl,alnefaie2023lkau23,zekiye2023aljawaab}, the shared task results revealed significant room for improvement in both tasks, particularly in handling the linguistic complexity and semantic nuances of Quranic texts.

This paper proposes a two-stage framework combining dataset expansion with tailored modeling: ensemble methods improve retrieval, while instruction-tuned LLMs with prompting enhance extraction.

Our main contributions are:
\begin{itemize}
    \item \textbf{Dataset Enhancement Strategy:} We expand the original dataset from 251 to over 2,000 questions through systematic paraphrasing and integration of diverse Quranic resources, significantly improving model training coverage and linguistic diversity.
    \item \textbf{Ensemble-Based Retrieval Framework:} We develop and evaluate multiple ensemble strategies for combining fine-tuned Arabic language models, achieving substantial improvements in passage retrieval accuracy.
    \item \textbf{Instruction-Tuned Extraction Approach:} We demonstrate that few-shot prompting with instruction-tuned large language models significantly outperforms fine-tuned approaches for answer extraction in low-resource settings.
    \item \textbf{State-of-the-Art Results:} Our framework achieves new benchmarks on the Quran QA 2023 Shared Task, with MAP@10 of 0.3128 and MRR@10 of 0.5763 for passage retrieval, and pAP@10 of 0.669 for answer extraction.
\end{itemize}

The remainder of this paper is organized as follows: Section II reviews related work in Arabic QA and Quranic text processing. Section III describes our methodology, including dataset enhancement, model architectures, and ensemble strategies. Section IV presents comprehensive experimental results and analysis. Section V discusses implications and limitations, and Section VI concludes with future research directions.

\section{Related Work}

This section surveys prior research relevant to our study on Quranic Question Answering (QA), covering Arabic QA systems, ensemble methods for retrieval, instruction-tuned language models for answer extraction, and specific efforts in Quranic QA. We highlight how these works inform our approach to tackling the linguistic and semantic challenges of Classical Arabic in the Qur’an.

\subsection{Arabic QA Datasets}

The field of Arabic Question Answering (QA) has advanced significantly, driven by the adoption of deep learning techniques and transformer-based models~\cite{alkhurayyif2023}. Datasets like Arabic-SQuAD, which rely on machine-translated questions, suffer from quality issues that limit model performance in specialized domains~\cite{alkhurayyif2023}.

To address this, recent efforts have introduced more robust and domain-specific resources. For example, the \textit{ArabicaQA} dataset~\cite{abdallah2024arabicaqa} contains over 89,000 answerable and 3,700 unanswerable questions derived from Arabic Wikipedia. These models are typically fine-tuned on domain-specific datasets such as the Quranic Passage Collection (QPC), the Arabic Reading Comprehension Dataset (ARCD), and newer large-scale corpora aimed at improving generalization in non-English QA.

\subsection{Arabic QA Systems}

Several Arabic QA systems have been evaluated on benchmarks such as Arabic-SQuAD and ARCD. Mozannar et al.~\cite{mozannar2019} reported F1 scores of 48\% on Arabic-SQuAD and 51\% on ARCD, highlighting the challenges in processing Arabic text even with advanced neural architectures.

In addition, early efforts focused on religious texts. Durachman and Adriani~\cite{durachman2019} built a system to extract answers from Indonesian Quran translations, achieving only 60\% accuracy. Similarly, Bhaskar et al.~\cite{bhaskar2019} proposed a rule-based multi-task QA system combining biomedical tasks using ensemble methods and NLTK-style pipelines, but lacked deep semantic modeling in general-purpose or Quranic Arabic.

\subsection{Ensemble Approaches}

Ensemble methods improve QA systems, as shown in MRQA 2019~\cite{mrqa2019} where combining BiDAF, QANet,and BERT enhanced eneralization. However, Quranic QA presents unique challenges since retrieving isolated verses (āyāt) risks unintended doctrinal weighting, as meaning emerges from interconnected passages within tafsīr traditions~\cite{saeed2005interpreting}. Our system retrieves thematic passages rather than individual verses, preserving coherence while maintaining ensemble benefits.

Similarly, ElKomy and Sarhan~\cite{elkomy2022} applied ensemble methods in Arabic QA using AraELECTRA, AraBERT, and other transformer models. Their approach achieved improved partial Reciprocal Rank (pRR) scores, reaching up to 65\% on the Quranic Reading Comprehension Dataset (QRCD).

ElKomy and Sarhan~\cite{elkomy2022} further advanced ensemble-based Arabic QA by augmenting datasets through translation and question reformulation. They fine-tuned models like AraBERT and QARiB, reporting MAP@10 of 0.34 and MRR of 0.52, with notable gains on QRCD.

\section{Methodology}

\subsection{Tasks Definition}

\subsubsection{Passage Retrieval}

The Passage Retrieval task aims to identify up to 10 Qur'anic passages relevant to a question in MSA. Some questions have no answer in the Qur'an, in which case no passages should be returned. The dataset is an extended version of AyaTECv1.2, containing 1,266 topical passages and 251 questions, including 37 zero-answer questions.

\subsubsection{Machine Reading Comprehension (MRC)}
The MRC task involves extracting answers from a given Qur'anic passage in response to a question in MSA. Each passage may contain single, multiple, or no answers. The dataset is an extended version of QRCDv1.2, comprising 1,562 question–passage pairs, including 62 with no answer.

\begin{figure*}[!t]
    \centering
    \begin{minipage}{0.42\textwidth}
        \centering
        \includegraphics[width=1.2\linewidth,keepaspectratio]{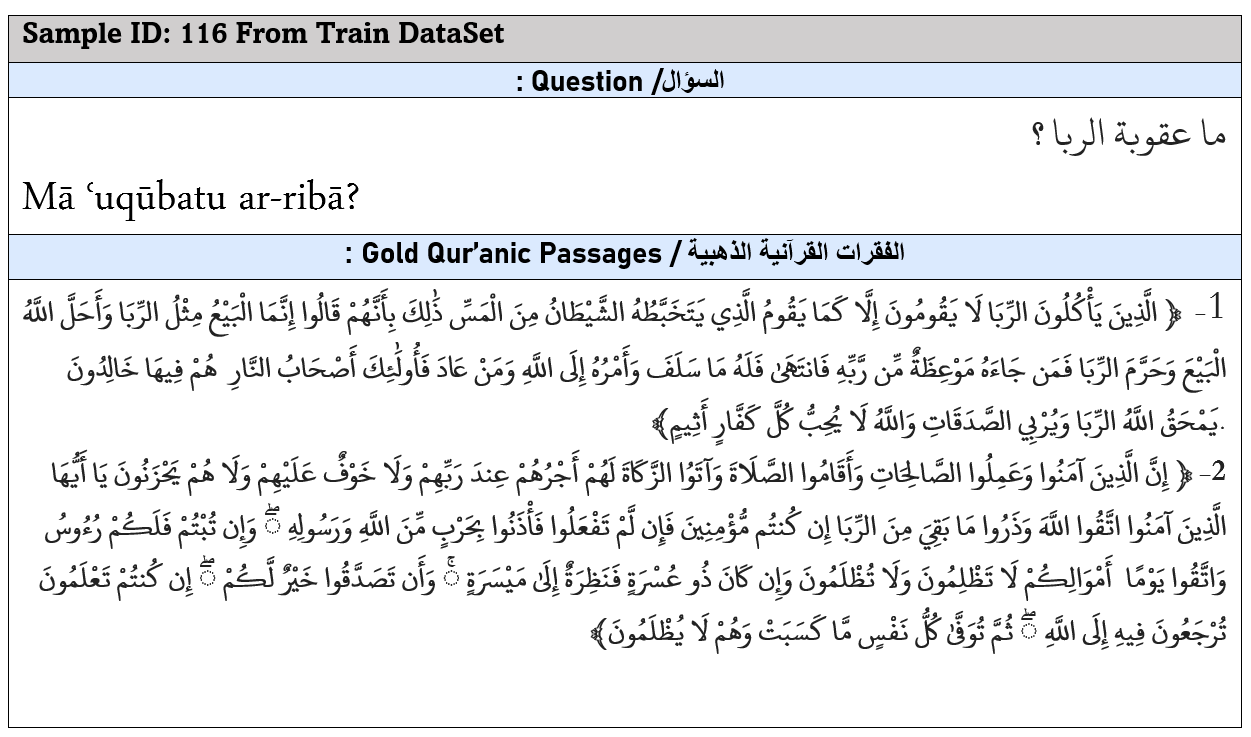}\\
        \small (a) Passage Retrieval Task
    \end{minipage}
    \hfill
    \begin{minipage}{0.42\textwidth}
        \centering
        \includegraphics[width=0.95\linewidth,keepaspectratio]{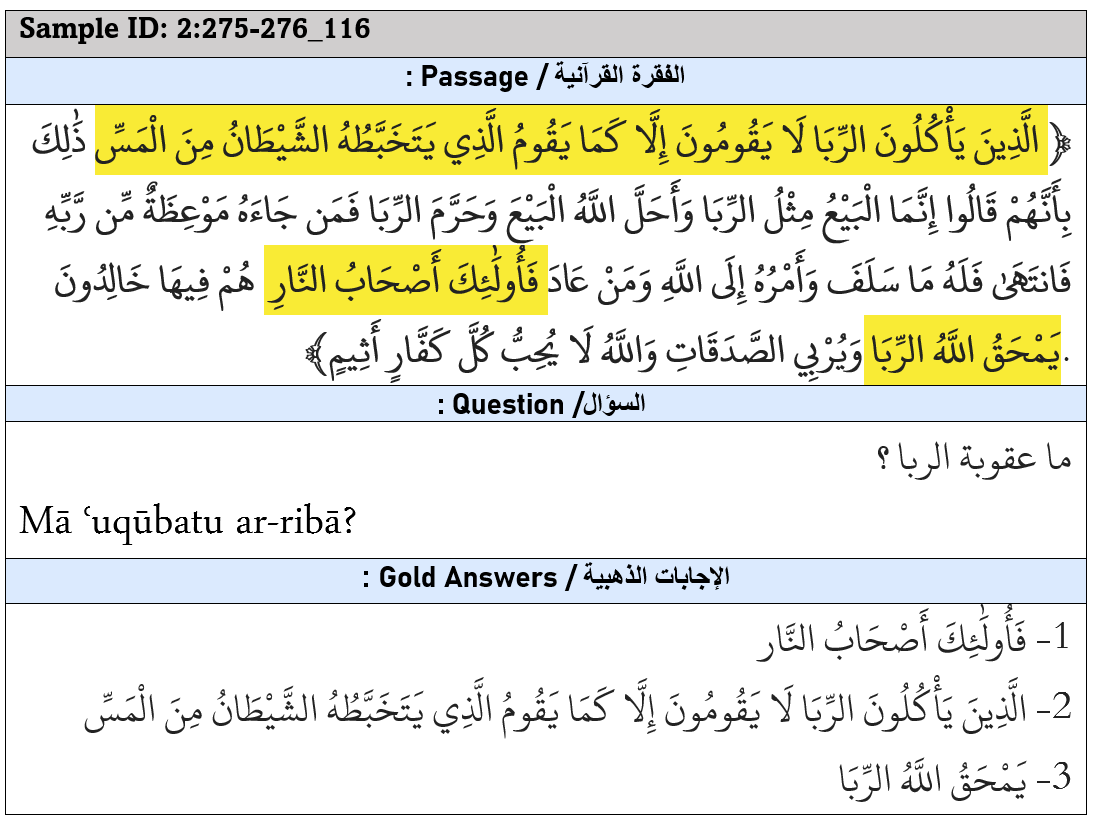}\\
        \small (b) Answer Extraction Task
    \end{minipage}
    \caption{Overview of retrieve the most relevant Qur’anic passage and extract the exact answer span.}
    \label{fig:tasks-graph}
\end{figure*}

\subsection{Dataset Collection}

\subsubsection{Passage Retrieval Task Dataset}

To improve coverage, diversity, and relevance of the Passage Retrieval task training set, we merged the official shared-task dataset with several external sources to form an expanded composite dataset. The combined dataset contains:

\begin{itemize}
\item \textbf{Quran QA 2023 Shared Task Dataset:} The official benchmark dataset containing \textbf{251 Arabic questions} linked explicitly to relevant passages within the Qur'anic Passage Collection (QPC). This dataset served as a Initial benchmarks and starting training examples for the model~\cite{malhas2020ayatec}.

\item \textbf{1,000 Questions and Answers from the Qur'an:}
A publicly available dataset compiled from Tafseer literature.
Its integration expanded our question-passage coverage, enabling the models to cover more Quranic subjects, linguistic differences and more in-depth analyses~\cite{ashor2023noor}.

\item \textbf{List of Plants Mentioned in the Qur'an and Hadith:}
A focused thematic dataset covering plant references in Islamic texts.
The resource has been utilized to enhance the capacity of the model to address questions regarding nature and Environmental issues addressed in the Qur'an~\cite{qbg2021plants}.

\item \textbf{Tafsir al-Jalalayn Pairs:}
A collection of Quranic verses, in pairs. with their Tafsir of Al-Jalalayn. This helped the model connect every verse with its interpretation, enhancing comprehension of significance and context~\cite{tafsir2021jalalayn}.

\item \textbf{TyDi QA (Arabic Subset):}  
A multilingual QA dataset with questions and answers in Arabic. It enhanced the performance of the model on general Arabic question answering~\cite{clark2020tydi}.
\end{itemize}

By integrating these diverse resources, we built a more
and more representative dataset. The addition enriched the ability of the model to generalize between common and specialized Quranic content, to aid better retrieval performance of \cite{basem2024optimized , OshallahCross}.

\subsubsection{Machine Reading Comprehension Task Dataset}

For MRC Task, we built a high-quality domain-specific dataset by merging and refining several complementary resources. Every resource was selected with care to
maximise linguistic diversity, thematic range, and structural coherence
consistency, maintaining high answer extraction performance.

\begin{itemize}
\item \textbf{QRCDv1.2 Dataset:} The official dataset released by Quran QA 2023, with fully annotated question-passage pairs. It was our \textbf{primary training and evaluation resource} type, including single-answer, multi-answer, and unanswerable questions~\cite{malhas2022arabic}.
    
\item \textbf{QUQA Dataset:} 
A varied set of questions pairs of passages sourced directly from the Qur'an. Its inclusion generated the dataset's thematic and linguistic diversity, enabling the models to deal with more semantic and Interpretative differences~\cite{quqa2023}.

\item \textbf{ARCD Dataset:} A general-purpose MSA reading comprehension set. We adapted it to better handle Arabic structure and improve Quranic communication.~\cite{arcd2021}.
\end{itemize}

For the convenience of standardizing all the datasets, we reformatted and standardized QUQA and ARCD to align with the QRCDv1.2
schema. This alignment enabled concerted training and evaluation workflows, regardless of the initial dataset structure.

The complete integration and normalization of our dataset
ensured that both prompt-based and fine-tuned models had
access to rich, diverse, and well-structured examples. This
significantly enhanced their capacity to apply knowledge to new and
unseen Quranic QA scenarios.

\begin{figure*}[!t]  
    \centering
    \includegraphics[width=0.96\textwidth]{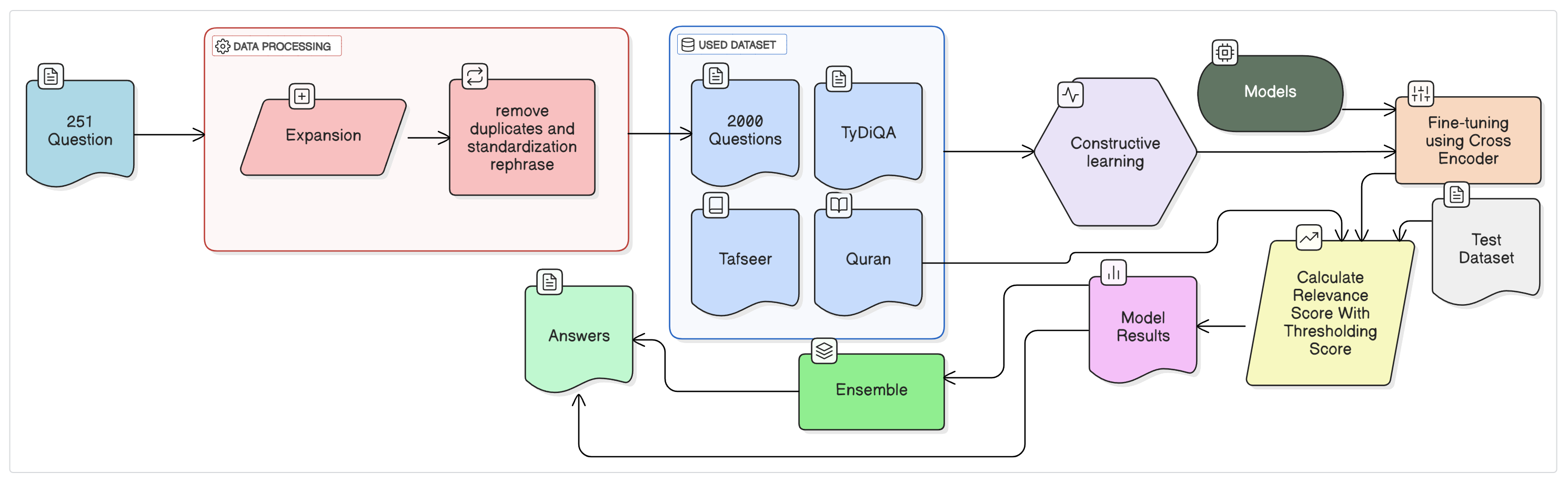}  
    \caption{Overview of Task A pipeline for Passage Retrieval, illustrating dataset expansion, integration of external resources (e.g., Tafseer, TyDiQA), model fine-tuning, relevance scoring via Cross-Encoder, and final ensemble-based ranking of Quranic passages.}
    \label{fig:TaskA}
\end{figure*}

\subsection{Dataset Expansion}

The initial Quran QA 2023 Shared Task dataset contained only \textbf{251 Arabic questions}, insufficient for training robust language models. We implemented a multi-phase expansion strategy to enhance both dataset size and linguistic diversity.

\subsubsection{Phase 1:  Expansion via External Sources}

We began by incorporating additional questions from external Quranic resources to extend the coverage beyond the original shared task dataset:

\begin{itemize}
    \item \textbf{1000 Questions and Answers from the Qur'an:} This resource provided a rich collection of question-passage pairs covering various Quranic themes. By selecting and curating relevant entries, we expanded the dataset from 251 to \textbf{665 questions}.
    
    \item \textbf{Plants in the Qur'an Resource:} In order to introduce more specialized content, we included a set of questions focusing on natural elements referenced in the Qur'an. This facilitated enhance the topical range and prepared the model for Addressing more specific inquiries from users.
\end{itemize}

\textbf{Phase 2 - Paraphrasing and Quality Review:} Each of the 665 questions was manually paraphrased into two alternative forms, creating three versions per question. All paraphrased questions underwent human annotation review for relevance and linguistic quality.

\textbf{Final Dataset:} The expansion process yielded 665 unique questions with 2,000 total variations, providing a robust foundation for retrieval model fine-tuning and few-shot prompt construction. This diversified dataset significantly enhanced system performance, particularly in the Passage Retrieval Task.

\subsection{Model Architecture}

\subsubsection{Task A: Passage Retrieval Architecture}
Task A employs uses a multi-step architecture designed to maximize passage retrieval for Quranic Question Answering. The method combines dataset augmentation, model fine-tuning, and ensemble techniques.

\paragraph{Data Expansion and Processing.}
The original 251-question dataset was expanded using external resources including Tafseer literature and thematic collections. Each question was paraphrased into two variants for linguistic diversity, yielding over 2,000 unique formulations. The dataset was then cleaned by removing duplicates and standardizing phrasing (Figure~\ref{fig:TaskA}).

\paragraph{Dataset Composition.}

The ultimate dataset merged
original shared task questions with outside questions
resources. In each question, a verse from the Quran was included , ensuring a comprehensive coverage of both general and domain-specific topics.

\paragraph{Model Fine-Tuning.}
Four pre-trained Arabic models were fine-tuned: \textbf{AraBERTv02-ARCD, AraELECTRA, CamelBERT-tydi-tafseer}, and \textbf{AraBERTv02-tydi-tafseer}. Training on the extended dataset improved \textbf{MAP@10} and \textbf{MRR@10} performance. Hyperparameter optimization explored learning rates, batch sizes, and sequence lengths to prevent overfitting and enhance generalizability.

\paragraph{Cross-Encoder Fine-Tuning and Relevance Scoring.}
After the initial model training, a Cross-Encoder was fine-tuned on the same data. This model was utilized to estimate pairwise relevance scores between every question and its candidate passages. A thresholding procedure was utilized to
Remove irrelevant results based on these scores.

\paragraph{Ensemble Strategy.}
An ensemble strategy was applied to merge the predictions of several fine-tuned models.
This strategy played to the strengths of every single model. The ensemble enhanced the robustness of the enhancements and improved the overall passage ranking accuracy.

\paragraph{Retrieval Performance.}
This architecture achieved competitive performance in the Quran QA 2023 Shared Task. The ensemble model achieved a MAP@10 score of \textbf{0.3128} and an MRR@10 of \textbf{0.5763}. The integration of dataset enlargement,model diversity, and ensemble methods led to the high effectiveness of the passage retrieval pipeline.

\subsubsection{Task B: Machine Reading Comprehension Architecture}

For mrc Task, we employed a comprehensive two-phase architecture tailored to span extraction challenges in low-resource contexts.

\paragraph{Phase 1: Fine-tuned Transformer Models}
Initially, we fine-tuned transformer-based Arabic models on QRCDv1.2, QUQA, and ARCD datasets. Models were trained to identify the start and end positions of answer spans within passages. However, their performance was constrained by limited exposure to varied linguistic structures.

\paragraph{Phase 2: API-Based Prompting with LLMs}
To overcome these limitations, we introduced API-based instruction-tuned large language models, namely:

\begin{itemize}
\item \textbf{Gemini API}
\item \textbf{DeepSeek API}
\end{itemize}

We generated three-shot prompts from examples in QRCDv1.2,
assisting the models in predicting correct spans. The results obtained from
these models were combined to make predictions more robust
and accuracy.

\paragraph{Final Span Extraction Output}
The merger of fine-tuned transformers and API models produced strong and
accurate extraction results. API-based models specifically outperformed transformers, and their total produced better consistency and enhanced overall accuracy.

\begin{figure}[ht]
    \centering
    \includegraphics[width=0.5\linewidth]{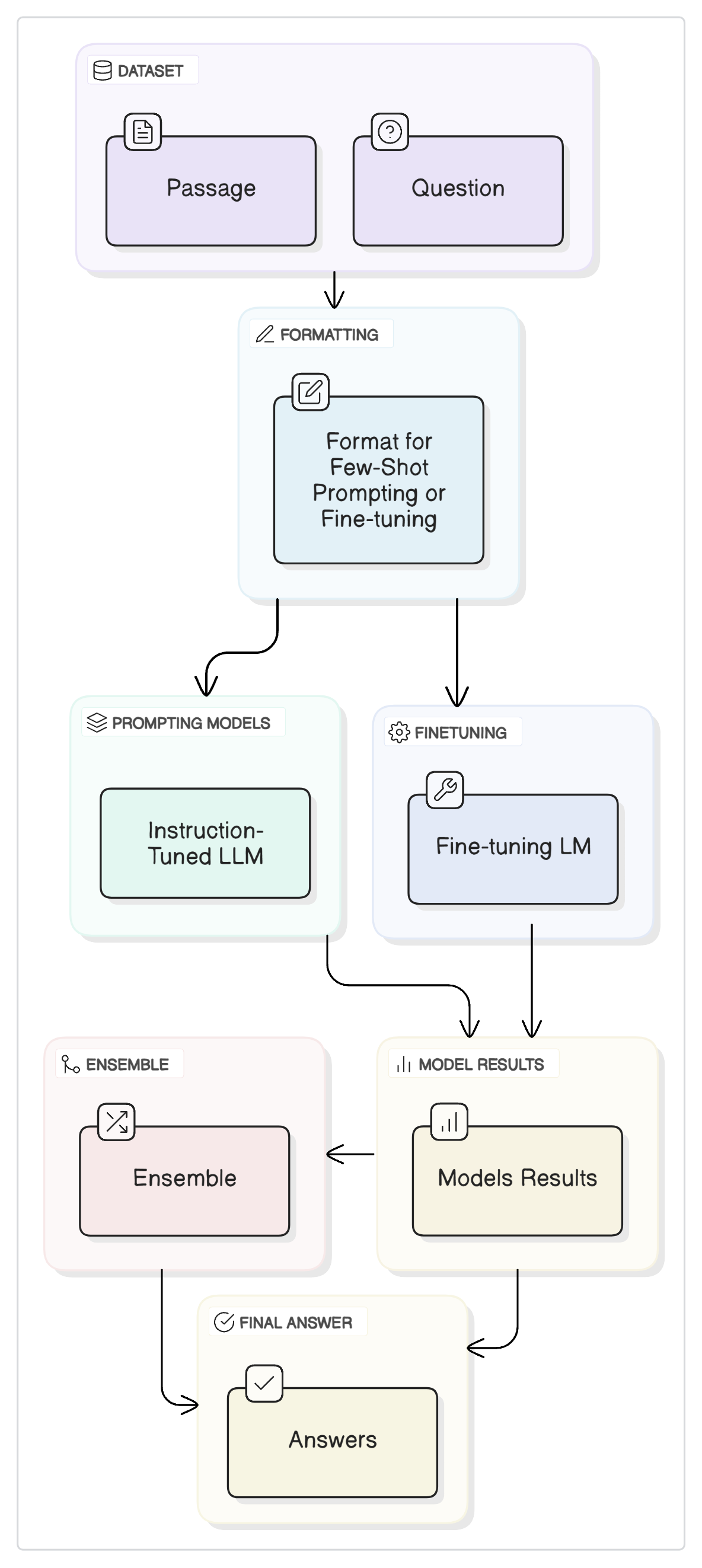}
    \caption{Overview of Answer Extraction task pipeline, covering data preparation, formatting for prompting or fine-tuning, model application (few-shot and fine-tuned), ensemble of model outputs, and final answer extraction.}
    \label{fig:enter-label}
\end{figure}

\subsection{Few-Shot Strategy and Prompt Engineering}\label{sec:fewshot}

To optimize the performance of API-based language models like \textit{Gemini} and \textit{DeepSeek}, we employed a \textbf{few-shot prompting} approach. This involves adding carefully selected examples to the prompt, helping the model understand the task \cite{basem2025few}.

Three instances from the QRCD v1.2 dataset were chosen to illustrate varied answering contexts: multiple correct spans, single correct span, and no solution. This ensures the model can handle diverse questions, from answerable to unanswerable.

We also designed a structured prompt to guide the model in generating accurate, text-aligned responses, minimizing unnecessary elaboration or hallucination. The prompt format includes clear instructions, examples, the input passage, and the question.

A sample of our prompt template is provided below:
\begin{itemize}
    \item \textbf{Answer-Extraction Task from Qur’anic Texts}
    
    \item \textbf{Instructions}:
    \begin{itemize}
        \item Extract all phrases from the text that answer the question, whether the answer is explicit or implicit.
        \item The answer must be a literal quotation from the passage.
        \item If no suitable answer is found, write: No answer found in the given text.
        \item Do not add any explanation or commentary.
        \item Consult the tafsīr (interpretation) of each verse to ensure accuracyto reach an accurate answer.
        \item Write the answers in the following format:
        \begin{itemize}
            \item ``First answer''
            \item ``Second answer''
        \end{itemize}
        \item \textbf{Examples}: \texttt{\{few\_shot\_text\}}
        \item \textbf{Answer the following passage}:
        \item \textbf{Passage}: \texttt{\{passage\}}
        \item \textbf{Question}: \texttt{\{question\}}
        \item \textbf{Answers}:
    \end{itemize}
\end{itemize}

The structured prompting method ensures precise, extractive responses while minimizing off-topic or fabricated outputs. It enhances model consistency when handling Quranic texts, which require exact phrasing and high semantic fidelity.

\begin{table*}[ht]
\centering
\caption{Performance Comparison for Passage Retrieval Task Before and After Dataset Extension, sorted in ascending order by MAP@10 after extension.}
\label{tab:task_a_comparison}
\begin{tabular}{l c c c c}
\toprule
\multirow{2}{*}{\textbf{Model}} & \multicolumn{2}{c}{\textbf{Before Extension}} & \multicolumn{2}{c}{\textbf{After Extension}} \\
\cmidrule{2-3} \cmidrule{4-5}
 & \textbf{MAP@10} & \textbf{MRR@10} & \textbf{MAP@10} & \textbf{MRR@10} \\
\midrule
AraBERTv02-ARCD & 0.2047 & 0.4445 & {0.2176} & 0.4709 \\
AraELECTRA & 0.2072 & 0.4473 & 0.2408 & 0.4934 \\
AraBERTv02-tydi-tafseer & 0.2494 & 0.5519 & 0.2673 & 0.5655 \\
CamelBERT-tydi-tafseer & 0.2508 & 0.5126 & {0.2926} & 0.4921 \\
{CamelBERT + AraBERTv02-ARCD + AraBERTv02-tydi} & N/A & N/A & \textbf{0.3009} & \textbf{0.5860} \\
CamelBERT + AraBERTv02-ARCD & N/A & N/A & 0.3017 & 0.5346 \\
\textbf{CamelBERT + AraBERTv02-tydi} & N/A & N/A & \textbf{0.3128} & \textbf{0.5763} \\
\bottomrule
\end{tabular}
\end{table*}

\subsection{Ensemble Method}

To enhance the overall performance of both Passage Retrieval Task and MRC Task, we employed ensemble techniques leveraging the strengths of multiple models.

\textbf{Passage Retrieval Task}
In this task, we employed two various ensemble techniques depending on model performance:

We ensembled AraBERTv02-ARCD and AraELECTRA predictions. The ensemble employed dynamic weighting by giving more importance to the predictions with confidence higher than 0.95. Predictions with more than 0.99 received an additional boost to prioritize highly confident outputs. We employed a weighted averaging strategy to merge rankings from both the models, which enhanced precision by prioritizing most consistent predictions without compromising on balanced ranking coverage.

We then employed Reciprocal Rank Fusion (RRF) to fuse predictions of three models: AraELECTRA, CamelBERT-tydi-tafseer, and AraBERTv02-tydi-tafseer. To maintain consistency among the results, we normalized the scores of all the models using min-max scaling. Lastly, the RRF approach was augmented with an exponential decay factor along with a confidence-based boosting mechanism specifically for scores above 0.8. Model-specific weights, determined based on prior performance evaluations, were applied, with AraELECTRA receiving the highest weight due to its superior individual performance. The final ranked list was generated by combining the RRF scores with the weighted normalized scores using a geometric mean, further improving retrieval accuracy across diverse question formulations.

\textbf{Machine Reading Comprehension Task}

In this task, we employed an ensemble multi-level strategy. We combined fine-tuned language models, specifically AraBERTv02 + CAMeLBERT and AraBERTv02 + BERT-large, for better extraction consistency among different model variants. This improved ensemble reached moderate improvements, its performance leveled off at pAP@10 scores of 0.517 and 0.509, respectively. 

For better performance, we further ensembled the top-performing API-based few-shot LLMs, namely Gemini (API-enhanced prompt) and DeepSeek (API-enhanced prompt). We merged their answer spans, keeping all unique responses while eliminating duplicates. This ensemble achieved the best overall performance with a pAP@10 of 0.669, outperforming all individual models and fine-tuned ensembles.

This layered ensemble approach—combining both fine-tuned models and instruction-tuned LLMs—demonstrated superior robustness and coverage, effectively handling both straightforward and ambiguous extraction scenarios.

\section{Experiments and Results}
\subsection{Evaluation Metrics}

For both Passage Retrieval task and MRC task, the Quran QA 2023 shared task uses standard evaluation metrics commonly employed in Information Retrieval (IR) and Question Answering (QA) tasks.

\textbf{Passage Retrieval}

The evaluation for this task is based on two primary metrics, Mean Average Precision at 10 (MAP@10) and Mean Reciprocal Rank at 10 (MRR@10):

\begin{itemize}
    \item \textbf{Mean Average Precision at 10 (MAP@10):} MAP@10 evaluates the average precision of relevant passages within the top 10 retrieved passages. It is calculated as follows:
    \[
    \text{MAP@10} = \frac{1}{|Q|}\sum_{q \in Q}\frac{1}{\min(R_q, 10)} \sum_{k=1}^{10}P(k)\,\delta_k
    \]
    where \(Q\) denotes the collection of inquiries,and \(R_q\) is the total relevant passages for query \(q\), \(P(k)\) is precision at rank \(k\), and \(\delta_k=1\) if the passage at rank \(k\) is relevant and 0 otherwise.

    \item \textbf{Mean Reciprocal Rank at 10 (MRR@10):} MRR@10 evaluates the rank position of the first relevant excerpt. It is defined as:
    \[
    \text{MRR@10} = \frac{1}{|Q|}\sum_{q \in Q}\frac{1}{\text{rank}_q}
    \]
    where \(\text{rank}_q\) is the rank of the first relevant passage retrieved within the top 10 positions for query \(q\). If no relevant passage is retrieved, the reciprocal rank is 0.
\end{itemize}

\textbf{Machine Reading Comprehension (MRC)}

For this task, the evaluation metric is partial Average Precision at 10 (pAP@10):

\begin{itemize}
    \item \textbf{partial Average Precision at 10 (pAP@10):} This metric calculates the precision of exact answer spans extracted from the top-ranked passages. pAP@10 considers partial correctness of the extracted spans and is calculated similarly to MAP, but with partial relevance scores:
    \[
    \text{pAP@10}(q) = \frac{1}{\min(R_q,10)}\sum_{k=1}^{10}P(k)\,\delta_k^{\text{partial}}
    \]
    Here, \(\delta_k^{\text{partial}}\) represents the partial correctness of the answer span at rank \(k\)is partially correct, with values from 0 to 1 depending on the overlap with the correct span.
\end{itemize}
All the measurements are obtained using the official evaluation script provided by the Quran QA 2023 shared task organizers.

\subsection{Experiment Comparison}

Table~\ref{tab:task_a_comparison} compares Passage Retrieval task results before and after dataset expansion, showing consistent improvements.

CamelBERT-tydi-tafseer achieved the best individual post-expansion performance with MAP@10 of 0.2926 and MRR@10 of 0.4921. An ensemble of CamelBERT, AraBERTv02, and AraBERTv02-tydi further improved results to MAP@10 of 0.3116 and MRR@10 of 0.5736.

Table~\ref{tab:our_models_summary} details MRC task performances of fine-tuned and few-shot LLM models. Few-shot LLMs with API-augmented prompting outperformed traditional fine-tuned models.

The best individual model was Gemini (API-enhanced prompt) led with a pAP@10 of 0.637, followed by DeepSeek at 0.624. Simple API usage without enhanced prompting scored lower (approximately 0.592--0.593), highlighting the importance of prompt optimization.

Fine-tuned models lagged behind, with AraBERTv02 (Merged DS - Tydi) top among them at 0.503. Model ensembles such as AraBERTv02 + CAMeLBERT and AraBERTv02 + BERT-large scored 0.517 and 0.509 respectively.

The highest overall performance was realized through enseming Gemini (API-enhanced prompt) and DeepSeek (API-enhanced prompt), with 0.669 pAP@10, a considerable improvement over the best single model. This confirms that
Combining complementary LLM predictions leads to superior results in extractive QA.

\begin{table}[ht]
\centering
\caption{Performance of our top models.}
\label{tab:our_models_summary}
\begin{tabular}{l l l}
\toprule
\textbf{Model} & \textbf{Type} & \textbf{pAP@10} \\
\midrule
\textbf{Gemini + DeepSeek (Ensemble)} & \textbf{Few-shot LLM} & \textbf{0.669} \\
{Gemini (API-enhanced prompt)} & {Few-shot LLM} & {0.637} \\
DeepSeek (API-enhanced prompt) & Few-shot LLM & 0.624 \\
DeepSeek (API) & Few-shot LLM & 0.593 \\
Gemini (API) & Few-shot LLM & 0.592 \\
{AraBERTv02 + CAMeLBERT (Ensemble)} & {Fine-tuned} & {0.517} \\
{AraBERTv02 + BERT-large (Ensemble)} & {Fine-tuned} & {0.509} \\
AraBERTv02 (Merged DS - Tydi) & Fine-tuned & 0.503 \\
BERT-large (Merged DS - Tydi) & Fine-tuned & 0.478 \\
CAMeLBERT (Merged DS - Tydi) & Fine-tuned & 0.442 \\
BERT-large (QRCDv1.2 - Tydi) & Fine-tuned & 0.438 \\
AraELECTRA (Merged DS - Tydi) & Fine-tuned & 0.425 \\
\bottomrule
\end{tabular}
\end{table}

\subsection{Results}
\label{sec:results}
Table ~\ref{tab:task_a_results} and Table ~\ref{tab:task_b_results} are representative of the official leaderboard The results of the Quran QA 2023 Shared Task for Passage Retrieval task and MRC task, respectively.
Our submissions got the highest scores on both tasks.
outdoing every other rival team and the established standards.

In \textbf{passage retrieval task}, our collaborative submission, \textbf{CamelBERT} +
\textbf{AraBERTv02-tydi} scored a MAP@10 of \textbf{0.3128} and
MRR@10 of \textbf{0.5763}, beating the next best competitor (TCE's run M00 with 0.2506 MAP@10 and 0.4610 MRR@10) by a significant margin.
Similarly,in \textbf{mrc task}, our ensemble of \textbf{Gemini and
DeepSeek}, using API-augmented prompting, attained a pAP@10.
of 0.669, outscoring the leading leaderboard submission by
TCE (0.5711 pAP@10).
The results explain the success of our method,
Complementing dataset augmentation, fine-tuning methods, and ensemble methods
techniques to maximize performance in both passages
retrieval and extractive span detection

\vspace{0.5em}
\begin{table}[ht]
\centering
\caption{Official Task A Leaderboard (MAP@10 and MRR@10).}
\label{tab:task_a_results}
\begin{tabular}{l l c c}
\toprule
\textbf{Team} & \textbf{Run} & \textbf{MAP@10} & \textbf{MRR@10} \\
\midrule
\textbf{Ours} & \textbf{Ensemble} & \textbf{0.3128} & \textbf{0.5763} \\
TCE & M00 & 0.2506 & 0.4610 \\
TCE & A00 & 0.2464 & 0.4940 \\
TCE & C00 & 0.2302 & 0.4706 \\
AHJL & SG2 & 0.1995 & 0.3889 \\
AHJL & SWOP3 & 0.1318 & 0.3021 \\
LKAU23 & run63 & 0.1242 & 0.3750 \\
AHJL & SS1 & 0.1202 & 0.2907 \\
LKAU23 & run61 & 0.1166 & 0.3632 \\
\textit{Baseline} & \textit{BM25} & \textit{0.0904} & \textit{0.2260} \\
\bottomrule
\end{tabular}
\end{table}

\vspace{0.5em}
\begin{table}[ht]
\centering
\caption{Official Task B Leaderboard (pAP@10).}
\label{tab:task_b_results}
\begin{tabular}{l l c}
\toprule
\textbf{Team} & \textbf{Run} & \textbf{pAP@10} \\
\midrule
\textbf{Ours} & \textbf{Gemini + DeepSeek (Ensemble)} & \textbf{0.669}\\
TCE & 4dfb8d601 & 0.5711 \\
TCE & dac0bdf4b & 0.5643\\
Al-Jawaab & tpgp4 & 0.5457 \\
Al-Jawaab & tgp4 & 0.5393 \\
TCE & ccc877dca & 0.5311 \\
LKAU23 & run03 & 0.5008 \\
LKAU23 & run02 & 0.4989\\
LowResContextQA & run01 & 0.4745 \\
LowResContextQA & run02 & 0.4745 \\
LowResContextQA & run03 & 0.4745 \\
GYM & run0 & 0.4613 \\
GYM & ensemble & 0.4588 \\
LKAU23 & run01 & 0.4541 \\
GYM & test1 & 0.4304 \\
\textit{Baseline} & \textit{WholePassage} & \textit{0.3268} \\
\bottomrule
\end{tabular}
\end{table}

\subsection{Discussion}

This paper discussed the principal issues in Quranic Question
Able to answer passage and span extraction questions.
Increasing the dataset from 251 to over 665 distinct questions, alongside paraphrasing, were essential to improving
model generality and retrieval accuracy. Although fine-tuning the models for mrc task had limited success because
due to data paucity, shifting to instruction-tuned API models
like \textbf{Gemini} and \textbf{DeepSeek}.
These performed better by
\textbf{few-shot prompt} without further training. \textbf{Ensembling} these
models continued to improve stability and performance throughout
both responsibilities. Nevertheless, the use of APIs presents
reproducibility and transparency concerns. Future work should focus on open, reproducible systems using graph-based linking and retrieval-augmented generation for deeper semantic reasoning across Quranic verses.

\section{Conclusion}
In this paper, we have suggested a \textbf{robust response system} to the Quranic question that achieves state-of-the-art results
for\textbf{ passage retrieval task} and \textbf{mrc task} in the \textbf{Quran QA 2023 Shared Task.}
In order to overcome the limited coverage and capacity of the initial
In first task, we increased the size of our training set considerably.
by leveraging several outside resources, manual question
generation and extensive rephrasing, leading to a diversified
dataset of over \textbf{2,000 questionnaires}. Using
on this enhanced dataset, we fine-tuned state-of-the-art Arabic language
models—\textbf{AraBERT}, \textbf{CamelBERT}, and \textbf{ELECTRA}—and 
used an \textbf{ensemble approach}, resulting in better MAP@10 and
MRR@10 rankings against the official winning outcomes.
For second task, recognizing the challenge of fine-tuning
small datasets, we successfully leveraged \textbf{instruction-tuned
Large Language Models (LLMs)}, namely \textbf{Gemini} and
\textbf{DeepSeek} operates in a few-shot prompting setting. By carefully
generating good prompts and corroborating findings by means of
With an ensemble method, our system achieved a pAP@10 score of
\textbf{0.669}, surpassing the top reported scores from the common
task.
Our research demonstrates that leveraging \textbf{dataset expansion, model ensembling, and few-shot prompting} with LLMs effectively advances the state-of-the-art of Arabic QA systems. This research not only sets a new benchmark for The Quranic QA also offers an explicit methodological framework.

Future work will enhance access to Islamic understanding by employing \textbf{Graph Neural Networks (GNNs)} to model deeper semantic links between passages, improving retrieval accuracy and context-aware answer extraction.

\section*{Acknowledgment}
This publication was supported by Qatar University High Impact Grant, Number LAW: 775. The findings achieved herein are solely the responsibility of the authors.

\bibliographystyle{IEEEtran}

\end{document}